# Multicriteria Group Decision-Making Under Uncertainty Using Interval Data and Cloud Models

Hadi A. Khorshidi and Uwe Aickelin

**Abstract**

In this study, we propose a multicriteria group decision-making (MCGDM) algorithm under uncertainty where data is collected as intervals. The proposed MCGDM algorithm aggregates the data, determines the optimal weights for criteria and ranks alternatives with no further input. The intervals give flexibility to experts in assessing alternatives against criteria and provide an opportunity to gain maximum information. We also propose a novel method to aggregate experts' judgements using cloud models. We introduce an experimental approach to check the validity of the aggregation method. After that, we use the aggregation method for an MCGDM problem. Here, we find the optimal weights for each criterion by proposing a bi-level optimisation model. Then, we extend the technique for order of preference by similarity to ideal solution (TOPSIS) for data based on cloud models to prioritise alternatives. As a result, the algorithm can gain information from decision-makers with different levels of uncertainty and examine alternatives with no more information from decision-makers. The proposed MCGDM algorithm is implemented on a case study of a cybersecurity problem to illustrate its feasibility and effectiveness. The results verify the robustness and validity of the proposed MCGDM using sensitivity analysis and comparison with other existing algorithms.

**Keywords:** Group Decision-Making, Cloud Model, Bi-level Optimisation, Interval-valued Data, Uncertainty.

## 1. Introduction

Decision making is an everyday task in human life and it becomes challenging when it includes multiple criteria [1, 2]. Multicriteria decision-making (MCDM) is an important area in decision-making theory [3]. Decisions ideally should not be made by individuals especially when the consequences affect a group of people. In addition, decision-making can often be improved by involving more than one expert [4]. The aim of multicriteria group decision-making (MCGDM) is to find the best option using preferences expressed by a set of decision-makers against a set of criteria [5, 6].

It normally is difficult for decision-makers to give absolutely certain values when examining the options. This difficulty comes from the nature of human thinking and the complication of the options and criteria [7]. Capturing uncertainty in decision-making has been getting much interest recently. There are two principal sources of uncertainty when collecting opinions 1) *inter-expert uncertainty* which is the variation among different decision-makers, 2) *intra-expert uncertainty* which comes from change of the mind of an individual decision-maker on the same situations [8].

One way to capture the opinions of decision-makers with their associated uncertainty are linguistic variables allowing computing with words [9]. This concept has attracted much interest and many techniques have been developed to deal with opinions which are expressed by words [10]. In these techniques, the words are typically encoded into quantity using triangular fuzzy numbers [11], generalized trapezoidal fuzzy numbers [12, 13], type-2 fuzzy sets [14], hesitant fuzzy sets [15] or cloud models [16, 17]. Then, quantitative analyses are applied, and the results are decoded into recommendations.

However, in these techniques, often limited linguistic terms are offered to decision-makers to express their opinion. So, there is still restricted flexibility for decision-makers in terms of capturing their uncertainty. This issue has been addressed as a limitation of studies using these techniques in [18], i.e. the linguistic terms are required to be limited and pre-defined, but providing more options may improve the performance of decision-making. In these techniques, each linguistic term is translated into a specific quantity. Whereas in reality, people may have different perceptions over a linguistic term in different situations [19]. Also, in an investigation on Analytic Hierarchy Process (AHP), researchers show that Fuzzy AHP by translating the 9-level evaluation into fuzzy number does not introduce much difference [20].

The proposed MCGDM is firstly motivated to overcome the limitations of studies using linguistic terms and give more flexibility to decision-makers to express their opinions. In this study, we use intervals to capture uncertainty in decision-makers' opinions. Using intervals, decision-makers can rate alternatives based on criteria flexibly and express their level of uncertainty to the precise extent they want. So, the proposed method does not give limited linguistic terms to decision-makers and does not need encoding of the linguistic terms into limited uncertain quantities. As a result, we gain maximum information from surveys.

In several studies that used interval-values for decision-making such as [8, 21], it is assumed that experts' ratings are uniformly distributed within the interval. However, in real world scenarios, experts do not have a similar level of uncertainty for the whole range of the interval. Moreover, these intervals are a better fit for uncertain data sourced from measurement errors and random observations. Intervals with Gaussian membership functions can be a better representation of experts' ratings. So the second motivation for the proposed MCGDM method is to capture uncertainty of the intervals using Gaussian membership functions. To address this motivation, we need to use cloud models to aggregate interval-valued data collected from experts.. Cloud models are introduced and have been extensively studied as a transformation model of uncertainty between the qualitative concept and the quantitative description [22]. In this study, we intend to investigate whether we can propose a method, using cloud models to aggregate interval-valued data, which is both precise in capturing information and relatively simply formulated using less computation time. Thus, the proposed method can provide an opportunity for a faster and more dynamic aggregation which is useful for many applications such as online ratings.

Another limitation of existing MCGDM algorithms is that the weights for criteria and/or decision-makers should be known [7, 23]. For many MCGDM problems, decision-makers need to give weights to criteria separately [3, 7, 23]. The third motivations for the proposed MCGDM is to assign the weights automatically through the aggregation process. Thus, we introduce a bi-level optimisation model which can derive the weights for criteria based on the interval-valued data collected from decision-makers. In our proposed MCGDM algorithm, decision-makers can express their opinions flexibly with different levels of uncertainty, and the alternatives are examined with no further information from decision-makers.

## 2. Preliminaries
### 2.1. Interval-valued Data

There are many situations in real life where the use of interval-valued data is more suitable than single crisp values. These situations occur due to different reasons. For instance, some phenomena cannot be expressed by a single value, e.g., the number of times that a patient smokes in a day. Therefore, we require some tolerance in measuring attributes especially in medical and engineering data. Moreover, often there are several observations for an attribute, e.g., testing blood sugar of a patient on different days or a group of people rating their online shopping's experience [24, 25].

Interval-valued data have been studied as an efficient representation for uncertain and imprecise information for different applications such as modelling survey data [21], prediction [26] and clustering [27].

There have been studies [14, 21] to aggregate interval-valued data into type-2 fuzzy sets. These studies assume that intervals have a uniform distribution. In addition, the aggregation process may require a large amount of computation and memory since the number of parameters is proportional to the number of surveyed intervals [28].

Here, we describe the aggregation method called interval agreement approach (IAA) [21]. Then, we use it in our comparative investigation. Suppose I=[a, b] is an interval and we have a set of intervals as $\mathcal{I} = \{I_1, I_2, \ldots, I_n\}$. The membership function of the set ($\mu_\mathcal{I}$) which aggregates intervals based on where they are in agreement is obtained using Eq. (1).

$$\mu_\mathcal{I} = \sum_{i=1}^{n} z_i / \left( \bigcup_{j_1=1}^{n-i+1} \bigcup_{j_2=j_1+1}^{n-i+2} \cdots \bigcup_{j_i=j_{i-1}+1}^{n} (I_{j_1} \cap \ldots \cap I_{j_i}) \right), \quad (1)$$

where $z_i = i/n$ is the degree of membership and is equal to one for the values in which all the intervals are in agreement.

To simplify Eq 1 for coding, let $E_\mathcal{I} = \{e_1, e_2, \ldots, e_{2n}\}$ be the set of the sorted values of endpoints, and $F_\mathcal{I} = \{f_1, f_2, \ldots, f_{2n}\}$ be a set of indicators where $f_i$ are +1 and -1 if $e_i$ is a lower and upper endpoints respectively [8]. So, the membership function can be obtained using Eq. (2).

$$\mu_\mathcal{I}(e) = \begin{cases} 1/n, & e = e_1 \\ \mu_\mathcal{I}(e_{i-1}) + f_i/n, & e = e_i \text{ and } i = 2, \ldots 2n \\ \mu_\mathcal{I}(e_i), & e_i \leq e < e_{i+1} \text{ and } i = 1, \ldots, 2n-1 \end{cases} \quad (2)$$

### 2.2. Cloud Models

A cloud model is a cognitive model to represent uncertainty. It contains the properties of probability distributions and fuzzy membership functions. The cloud model reflects the correlation between randomness and fuzziness to construct a map between quality and quantity. The distribution of membership for a cloud model within the domain is named as a membership cloud or simply a cloud. The cloud is composed of cloud drops, in a way that the higher certainty degree of a cloud drop indicates that the cloud drop has higher probability of appearance [22, 23].

There are similarities between cloud models and type-2 fuzzy sets. They both have been developed to add a higher level of uncertainty to membership functions. However, they are different in terms of fundamental theory and motivations to represent the uncertainty. Type-2 fuzzy sets capture the uncertainty of membership functions by defining upper and lower membership functions and use analytic mathematical methods to analyse the uncertainty. Cloud models use discrete cloud drops to express the uncertainty motivated by the random distribution of samples, without defining a membership function. Cloud models can effectively integrate statistics with fuzzy logic (you can refer to [29] for more details on a comparative investigation between these two).

The advantage of cloud models is that a cloud can be characterized by only three parameters, which means low computation and memory requirements, while type-2 fuzzy sets have higher computational complexity [28, 29]. These three parameters are expectation ($Ex$), entropy ($En$) and hyper-entropy ($He$). $Ex$ is the centre value of the cloud drops in a cloud model. In other words, it is the representative of the cloud model with the highest probability of appearance. $En$ measures the uncertainty and represents the domain of the cloud. $He$ quantifies the dispersion of cloud drops and the randomness of the membership. In other words, it is the uncertainty degree of the entropy [23, 30].

Normal cloud models are based on a normal distribution (probability theory) and a Gaussian membership function (fuzzy logic) [30]. Normal cloud models have been used successfully in different applications such as knowledge discovery [31], image processing [32], decision making [33], food industry [34], risk evaluation [23].

*Definition 1* [22]: Let $U$ be the universe of discourse and $T$ be a concept in $U$. If $x \in U$ is a random instance of concept $T$, which satisfies $x \sim N\left(Ex, En^{'2}\right)$, $En^{'} \sim N(En, He^2)$ and certainty degree that $x$ belongs to concept $T$ is calculated by Eq. (3).

$$y = e^{-\frac{(x-Ex)^2}{2En^{'2}}}, \quad (3)$$

So, a cloud drop can be represented as $(x, y)$ and the normal cloud would be $\tilde{y} = (Ex, En, He)$. Once the three parameters of a normal cloud model are determined, we can generate the normal cloud [30] using Algorithm 1.

---
**Algorithm 1:** Normal cloud generator (CG)
---
**Input:** Parameters $Ex$, $En$, $He$ and number of cloud drops $N$
**Output:** $N$ cloud drops and their certainty degree.
1. **for** $i$ from 1 to $N$ **do**
2.    $En'_i \leftarrow$ generate a random value which follows a Normal distribution with mean $En$ and variance $He^2$
3.    $x_i \leftarrow$ generate a random value which follows a Normal distribution with mean $Ex$ and variance $En'^2$
4.    $y_i \leftarrow$ calculate the certainty degree /* Using Eq. (3) */
5. **end for**
---

Fig. 1 shows a normal cloud and its numerical characteristics which is generated by $(0, 1, 0.1)$ and 5000 cloud drops.

[Figure 1]

*Definition 2* [23, 35]: Suppose a normal cloud $\tilde{y} = (Ex, En, He)$ in the domain of $U$ and a positive non-zero value $\lambda$, then their multiplication is as Eq. (4).

$$\lambda \tilde{y} = (\lambda Ex, \sqrt{\lambda} En, \sqrt{\lambda} He), \qquad (4)$$

*Definition 3* [16, 23]: Given two normal clouds $\tilde{y}_1 = (Ex_1, En_1, He_1)$ and $\tilde{y}_2 = (Ex_2, En_2, He_2)$ in the domain of $U$, these two clouds can be compared through as follows.

- Convert the clouds into interval-values as $\tilde{y}_1 \approx [\underline{a}, \bar{a}]$ and $\tilde{y}_2 \approx [\underline{b}, \bar{b}]$, where $\underline{a} = Ex_1 - 3En_1$, $\bar{a} = Ex_1 + 3En_1$, $\underline{b} = Ex_2 - 3En_2$, $\bar{b} = Ex_2 + 3En_2$.
- Calculate $S_{\tilde{y}_1, \tilde{y}_2}$ using Eq. (5).
$$S_{\tilde{y}_1, \tilde{y}_2} = 2(\bar{a} - \underline{b}) - (\bar{a} - \underline{a} + \bar{b} - \underline{b}), \qquad (5)$$
- Use the following rules for the comparison.
  1. If $S_{\tilde{y}_1, \tilde{y}_2} > 0$, then $\tilde{y}_1 > \tilde{y}_2$;
  2. If $S_{\tilde{y}_1, \tilde{y}_2} = 0$ and $En_1 < En_2$, then $\tilde{y}_1 > \tilde{y}_2$;
  3. If $S_{\tilde{y}_1, \tilde{y}_2} = 0$, $En_1 = En_2$ and $He_1 < He_2$, then $\tilde{y}_1 > \tilde{y}_2$;
  4. If $S_{\tilde{y}_1, \tilde{y}_2} = 0$, $En_1 = En_2$ and $He_1 = He_2$, then $\tilde{y}_1 = \tilde{y}_2$.

*Definition 4*: The distance between two normal clouds $\tilde{y}_1 = (Ex_1, En_1, He_1)$ and $\tilde{y}_2 = (Ex_2, En_2, He_2)$ is measured as Eq. (6).

$$d(\tilde{y}_1, \tilde{y}_2) = \sqrt{|Ex_1 - Ex_2|^2 + |En_1 - En_2| + |He_1 - He_2|}, \qquad (6)$$

The distance measure should meet the following conditions [7].

1. $d(\tilde{y}_1, \tilde{y}_2) \geq 0$;
2. $d(\tilde{y}_1, \tilde{y}_2) = 0$ if and only if $\tilde{y}_1 = \tilde{y}_2$;
3. $d(\tilde{y}_1, \tilde{y}_2) = d(\tilde{y}_2, \tilde{y}_1)$;
4. For any normal cloud $\tilde{y}_3 = (Ex_3, En_3, He_3)$ in $U$, $d(\tilde{y}_1, \tilde{y}_2) \leq d(\tilde{y}_1, \tilde{y}_3) + d(\tilde{y}_3, \tilde{y}_2)$.

*Proof*: Clearly, the proposed distance measure in Eq. (6) meets the first three conditions. We prove that the proposed distance measure also meets the fourth condition as follows.

$d(\tilde{y}_1, \tilde{y}_2) = \sqrt{|Ex_1 - Ex_2|^2 + |En_1 - En_2| + |He_1 - He_2|} =$
$\sqrt{|Ex_1 - Ex_3 + Ex_3 - Ex_2|^2 + |En_1 - En_3 + En_3 - En_2| + |He_1 - He_3 + He_3 - He_2|} \leq$
$\sqrt{(|Ex_1 - Ex_3| + |Ex_3 - Ex_2|)^2 + |En_1 - En_3| + |En_3 - En_2| + |He_1 - He_3| + |He_3 - He_2|} \leq$
$\sqrt{|Ex_1 - Ex_3|^2 + |En_1 - En_3| + |He_1 - He_3| + |Ex_3 - Ex_2|^2 + |En_3 - En_2| + |He_3 - He_2|} \leq$
$\sqrt{|Ex_1 - Ex_3|^2 + |En_1 - En_3| + |He_1 - He_3|} +$
$\sqrt{|Ex_3 - Ex_2|^2 + |En_3 - En_2| + |He_3 - He_2|} = d(\tilde{y}_1, \tilde{y}_3) + d(\tilde{y}_3, \tilde{y}_2)$

## 3. Aggregation

We use interval values to collect the decision-makers' opinions. These values give decision-makers flexibility to express their opinions, i.e., allow for a degree of uncertainty. Once the values are collected, we need an aggregation method to create a formal representative of the opinions and keep all the information. Here we use a cloud model to aggregate the interval values. This aggregation method can be used for both inter-expert and intra-expert uncertainties. The advantage of using cloud models is to provide a simpler representation for uncertainty (just with three parameters). So, less computations and memory are needed.

We assume that each interval-value has a Gaussian membership function which we believe is a better assumption in comparison with the uniformly distributed membership function. Suppose we have an interval-value such as I=[a, b]. This value can be translated into a 99.73% confidence interval (CI) based on "$3\sigma$ principle" of the Normal distribution using Eq. (7).

$$\begin{cases} \bar{x} = (a+b)/2 \\ \sigma = (b-a)/6 \end{cases} \qquad (7)$$

where $\bar{x}$ is the mean and $\sigma$ is the standard deviation. So, the interval-value of I can be converted into a pair as $(\bar{x}, \sigma)$ which $\bar{x}$ is an estimation for the interval and $\sigma$ quantifies the confidence (uncertainty) of $\bar{x}$ estimate.

*Example 1:* Suppose we have three decision-makers, named as DM1, DM2 and DM3 respectively. They give their opinion using interval-values as DM1=[3,4], DM2=[1,6] and DM3=[2,5]. The probability density functions (pdf) for these opinions after translation into Gaussian membership function are visualized in Fig. 2. As can be seen, the more the certain opinion, the smaller the range and the higher the density peak.

[Figure 2]

Once the interval-values are translated, we use a cloud model to aggregate opinions. The parameters for the aggregated cloud model can be calculated as Eq. (8).

$$\begin{cases} Ex = \frac{1}{K}\sum_k \bar{x}_k \\ En = \frac{1}{K}\sum_k \sigma_k + \sqrt{\frac{1}{K}\sum_k (\bar{x}_k - Ex)^2} \\ He = \sqrt{\frac{1}{K}\sum_k (\sigma_k - En)^2} \end{cases} \qquad (8)$$

where *Ex*, *En* and *He* are the expectation, entropy and hyper-entropy of the aggregated cloud model respectively, and *K* is the number of decision-makers. The expectation of the aggregated cloud is calculated using the average of the estimation values of the intervals (Eq. 7) to represent the centre values of the cloud drops. The entropy value comes from two sources of uncertainties. First, the uncertainties of intervals obtained by Eq. (7). Second, the variation of intervals' estimations from the centre of the cloud model. As a result, the aggregated cloud model can cover the whole entropy in the aggregation process. The hyper-entropy calculates how dispersed uncertainties are across decision-makers.

For example 1, we can aggregate decision-makers' judgement using Eq. (8). The aggregation finds a cloud model with *Ex*=3.5, *En*=0.5 and *He*=0.27. Fig. 3 visualizes the output of the aggregation process. To achieve this, we firstly calculated probability values of each decision-maker based on their normal distribution using $e^{-\frac{(x-\bar{x})^2}{2\sigma^2}}$ (colored lines). Then, we generated 250 cloud drops based on the aggregated cloud model (black circles). As can be seen, the generated drops are clustered around the areas where decision-makers are in agreement as well as covering the uncertainty of decision-makers.

[Figure 3]

We also created more examples to show how the aggregation method works for different situations. The details for these examples and the aggregation results are shown in Table 1, and they are visualized in Fig. 4.

[Table 1]

In example 2, the cloud drops are concentrated around the middle point and scattered across the domain based on decision-makers' uncertainty. In example 3, the hyper-entropy is zero and the cloud model is a normal distribution as all decision-makers give the same intervals. The uncertainty of decision-makers in examples 4 and 5 is the same. However, the aggregated entropy and hyper-entropy of example 5 is lower due to the overlap between intervals (decision-makers are more in agreement). Examples 6 and 7 are instances where decision-makers have different estimations and different uncertainty levels. So, they have high aggregated entropy and hyper-entropy values. However, the cloud model in example 7 has a lower hyper-entropy as two of decision-makers are more in agreement.

[Figure 4]

### 3.1. Experiments

To validate the performance of the proposed aggregation method, we design some appropriate experiments. In these experiments, we simulate different situations where the number of decision-makers (*d*) varies from 2 to 10. For each situation, we generate *P* aggregation problems. In each problem, *d* interval-values are generated randomly. Firstly, we estimate the parameters of the Gaussian membership function equivalent with each interval-value using Eq. (7). Then, we generate *R* random values using the parameters for each distribution. We create a pool of randomly generated values and calculate the descriptive statistics such as mean, quartile 1, median (quartile 2) and quartile 3 for the pool. Secondly, we aggregated the interval-values into a cloud model using Eq. (8). Based on the aggregated parameters, we generate $R \times d$ random values, and similarly find means and quartile values. Finally, we calculate the similarity between the two inter-quartile ranges (IQR) using a similarity measure based on overlapping ratios ($S_{OR}$) [36] formulated as Eq. (9). Note that IQR denotes the range between quartiles 1 and 3.

$$S_{OR}(IQR_p, IQR_c) = \min\left(\frac{|IQR_p \cap IQR_c|}{|IQR_p|}, \frac{|IQR_p \cap IQR_c|}{|IQR_c|}\right), \quad (9)$$

where $|IQR_p \cap IQR_c|$ is the size of the intersection between two IQRs, $|IQR_p|$ and $|IQR_c|$ are the size of the IQR of the pool and the cloud respectively. The higher the value of $S_{OR}$, the more the similarity of two IQRs. The steps to implement the experiments are outlined in Algorithm 2.

---
**Algorithm 2:** Aggregation experiment
---
**Input:** Constant values for *D*, *P* and *R*
**Output:** $D - 1$ matrices ($O_d$) includes quartiles and $S_{OR}$

1. $D \leftarrow$ Upper bound for number of decision-makers; $P \leftarrow$ Number of problems; $R \leftarrow$ Number of random values
2. **for** *d* from 2 to *D* **do**
3.   $O_d \leftarrow$ An empty matrix with *P* rows and 9 columns
4.   **for** *p* from 1 to *P* **do**
5.     **for** *q* from 1 to *d* **do**
6.       $I_q \leftarrow$ Generate an interval-value randomly
7.       $Norm_{I_q} \leftarrow$ Find Gaussian membership function for $I_q$
8.       $Pool \leftarrow$ Generate *R* random values using $Norm_{I_q}$
9.     **end for**
10.     $O_d \leftarrow$ Find mean and quartiles of *Pool*

11.     Aggregate the interval-values
12.     $Cloud$ ← Generate $R \times d$ random values using the aggregated cloud model /* Using Algorithm 1 */
13.     $O_d$ ← Find mean and quartiles of $Cloud$
14.     $O_d$ ← Calculate $S_{OR}$ /* Using Eq. (9) */
15.   **end for**
16. **end for**

For this experiment, we set the inputs as $D = 10$, $P = 100$ and $R = 50$. Once we have the results, we test the difference between the mean values of the pool and the aggregated cloud using paired t-tests. We checked the normality of data before applying this parametric test. The outputs of statistical tests are shown in Table 2.

[Table 2]

The statistical test results show that we cannot reject the null hypothesis (true difference in mean is equal to zero) for mean and quartile 3 in all cases as p-values are greater than 0.05. This proves that the aggregation method can achieve similar mean values. Fig. 5 represents the results of the calculation of $S_{OR}$ for each problem within each situation. The red dot denotes the mean of $S_{OR}$ and the red line shows an interval width of two standard deviations. We can see the performance of the aggregation method is robust by changing the number of decision-makers.

[Figure 5]

We also investigate the performance of the aggregation method through a simulation approach. The parameters of a cloud model can be derived from cloud drops in the cloud. This transformation from the cloud drops to the cloud's parameters is called backwards cloud generator ($CG^{-1}$). There are two types of algorithm for backwards cloud generation for situations where the certainty degree is known [22]. However, we do not know the degree of certainty and hence require a different approach. The backwards cloud generator without knowing the certainty degree is described in Algorithm 3.

**Algorithm 3:** Backwards normal cloud generator ($CG^{-1}$)
**Input:** $x_i$ values of $N$ cloud drops
**Output:** Parameters $Ex$, $En$, $He$ of the cloud model
1.   $Ex$ ← calculate the mean of $x_i$
2.   $S^2$ ← calculate the variance of $x_i$
3.   $En$ ← calculate $\sqrt{\frac{\pi}{2}} \times \frac{1}{N}\sum_{i=1}^{N}|x_i - Ex|$
4.   $He$ ← calculate $\sqrt{|S^2 - En^2|}$

In other words, we use the generated pool (in Algorithm 2, steps 3-9) as the input for Algorithm 3 and derive the cloud's parameters. So, we obtain the cloud parameters using simulation. However, we need much more computation and memory. Consequently, we can obtain the quartiles and calculate $S_{OR}$ from Algorithm 2. Fig. 6 compares the average of $S_{OR}$ calculated within for each scenario.

[Figure 6]

As can be seen from Fig. 6, both the proposed aggregation method and the backwards cloud generator can retain a high percentage of information from the pool. This shows that using cloud models is a suitable way to aggregate information from interval-valued data. In addition, the results show that the proposed aggregation method performs better than the $CG^{-1}$ in almost all situations (except when number of decision-makers is 5). Moreover, the proposed aggregation method can simply estimate the parameters of the cloud model from the interval-valued data using Eq. (8), while the $CG^{-1}$ needs a pool of cloud drops to derive the parameters. Overall, the proposed aggregation method performs well in capturing uncertainty in the opinions of decision-makers, which is very helpful in MCGDM.

In addition, we compare the performance of the proposed aggregation method with other non-cloud models. Thus, we can ascertain how well the proposed method can gain information from intervals having Gaussian membership functions. We perform the comparison with two other methods. For the first alternative, we aggregate the intervals into an interval with Gaussian membership function by averaging the mean and standard deviation values. In other words, the intervals are aggregated into a type-1 fuzzy set. So, this aggregation does not consider the uncertainty of membership functions. The second method is IAA [21] which is described in section 2.1.

[Figure 7]

The comparison results are presented in Fig. 7. The results show that our proposed aggregation method performs better than two other methods by resulting in higher similarity values with generated inputs. Especially, it performs much better than type-1 fuzzy set aggregation. The results confirm that the proposed aggregation method can capture the uncertainty across the membership functions well.

## 4. Proposed MCGDM

In this section, a decision-making model using the proposed aggregation method (from section 3) is developed to prioritise alternatives when criteria are assessed under uncertainty by a group of decision-makers using interval values. For a MCGDM problem, suppose there are $K$ decision-makers $DM_k (k = 1,2,...,K)$, $M$ criteria $C_j (j = 1,2,...,M)$ and $N$ alternatives $A_i (i = 1,2,...,N)$.

The proposed MCGDM model consists of three main stages. The first stage includes collecting uncertain data from decision-makers and aggregating decision-makers' opinions. In the second stage, the decision matrix is constructed for alternatives and criteria. This is followed by weights for each criterion being determined by an optimisation model. In the third stage, the alternatives are prioritised using an extended TOPSIS method. Each stage consists of detailed steps which are described as follows.

### Stage 1: data collection and aggregation

*1) Evaluate alternatives using interval-values*

Due to the complexity and vagueness of human thinking, it is hard for decision-makers to express their opinions without uncertainty. To give more flexibility to decision-makers, we allow the use of intervals to evaluate alternatives based on different criteria. Intervals provide an opportunity for decision-makers to express their opinions with different levels of uncertainty.

*2) Aggregate the opinions of decision-makers*

The interval-values are aggregated and translated into cloud models using Eq. (8).

### Stage 2: decision matrix construction

*1) Construct the decision cloud matrix*

After aggregation, the information gained from decision-makers is translated into cloud models. For each criterion of each alternative, there will be a cloud model denoted by $\tilde{y}_{ij} = (Ex_{ij}, En_{ij}, He_{ij})$. Then, the decision matrix $\tilde{Y}$ can be constructed as Eq. (10).

$$\tilde{Y} = \begin{matrix} & C_1 & C_2 & \cdots & C_M \\ A_1 \\ A_2 \\ \vdots \\ A_N \end{matrix} \begin{bmatrix} \tilde{y}_{11} & \tilde{y}_{12} & \cdots & \tilde{y}_{1M} \\ \tilde{y}_{21} & \tilde{y}_{22} & \cdots & \tilde{y}_{2M} \\ \vdots & \vdots & \ddots & \vdots \\ \tilde{y}_{N1} & \tilde{y}_{N2} & \cdots & \tilde{y}_{NM} \end{bmatrix}, \quad (10)$$

*2) Determine the weights of criteria*

Once the decision cloud matrix is constructed, we can use the hyper-entropy of cloud models to give weights to criteria. The idea is to give less weight to the criteria with higher hyper-entropy (dispersion) across all alternatives. In other words, the criteria where decision-makers have higher level of agreement, are more important in ranking alternatives. This is because those criteria are more reliable to build the decision-making process than the criteria on which the collective opinion of the decision-makers has high uncertainty. These weights help the decision-making process becomes more robust and less sensitive to dispersed opinions of the decision-makers. In addition, finding the level of agreement of decision-makers has always been important in decision-making processes [19]. We use a bi-level optimisation model to find the weights of the criteria as Eq. (11). The optimal weights are obtained by considering all alternatives in one model.

$$\min \max_{ij} \{|He_{ij} w_j - He_{ij_o} w_{j_o}|\} \quad (11)$$

Subject to

$\sum_j w_j = 1$

$w_j \geq 0$, for all $j$

where $w_j$ is the weight for $j$th criterion and $j_o$ refers to one of the criteria. This bi-level optimisation model can be transformed into a linear programming problem using the following formulations.

$$\min \xi \quad (12)$$

Subject to

$He_{ij} w_j - He_{ij_o} w_{j_o} \leq \xi$, for all $i$ and $j \neq j_o$

$He_{ij} w_j - He_{ij_o} w_{j_o} \geq -\xi$, for all $i$ and $j \neq j_o$

$\sum_j w_j = 1$

$w_j \geq 0$, for all $j$

The optimal weights for criteria are obtained by solving problem in Eq. (12). We use the optimal weights to update the decision cloud matrix.

*3) Construct the weighted decision cloud matrix*

The weighted decision cloud matrix $\hat{\tilde{Y}}$ is constructed by multiplying the elements of the decision cloud matrix $\tilde{Y}$ with the weights of criteria as Eq. (13).

$$\hat{\tilde{y}}_{ij} = \tilde{y}_{ij} \times w_j, \text{ for all } i \text{ and } j \quad (13)$$

where $\hat{\tilde{y}}_{ij}$ are the elements of the weighted decision cloud matrix, which is constructed as Eq. (14).

$$\hat{\tilde{Y}} = \begin{matrix} & C_1 & C_2 & \cdots & C_M \\ A_1 \\ A_2 \\ \vdots \\ A_N \end{matrix} \begin{bmatrix} \hat{\tilde{y}}_{11} & \hat{\tilde{y}}_{12} & \cdots & \hat{\tilde{y}}_{1M} \\ \hat{\tilde{y}}_{21} & \hat{\tilde{y}}_{22} & \cdots & \hat{\tilde{y}}_{2M} \\ \vdots & \vdots & \ddots & \vdots \\ \hat{\tilde{y}}_{N1} & \hat{\tilde{y}}_{N2} & \cdots & \hat{\tilde{y}}_{NM} \end{bmatrix}, \quad (14)$$

***Stage 3: alternative prioritization***

Once the weighted decision cloud matrix is available, we can use the extended TOPSIS method to rank alternatives. The extended TOPSIS can deal with cloud models as its values.

*1) Define ideal solutions*

In TOPSIS, we need to define positive and negative ideal solutions. The positive (negative) ideal solution is the best (worst) solution from all alternatives. To define these ideal solutions, we should check whether a criterion is a higher-the-better or lower-the-better one. Then, the ideal solutions can be obtained by Eqs. (15) and (16).

$A^+ = \{\max_i \hat{\tilde{y}}_{ij} : j \in J, \min_i \hat{\tilde{y}}_{ij} : j \in J' \mid i = 1,2,\ldots,N\} = \{\hat{\tilde{y}}_1^+, \hat{\tilde{y}}_2^+, \ldots, \hat{\tilde{y}}_M^+\},$ (15)

$A^- = \{\min_i \hat{\tilde{y}}_{ij} : j \in J, \max_i \hat{\tilde{y}}_{ij} : j \in J' \mid i = 1,2,\ldots,N\} = \{\hat{\tilde{y}}_1^-, \hat{\tilde{y}}_2^-, \ldots, \hat{\tilde{y}}_M^-\},$ (16)

where $A^+$ and $A^-$ are the positive and negative ideal solutions respectively, $J$ refers to the set of the higher-the-better criteria and $J'$ refers to the set of the lower-the-better criteria.

*2) Calculate distance from ideal solutions*

The distance between each alternative and the ideal solutions can be calculated using Eqs. (17) and (18).

$d_i^+ = \sum_j d(\hat{\tilde{y}}_{ij}, \hat{\tilde{y}}_j^+)$, for all $i$ (17)

$d_i^- = \sum_j d(\hat{\tilde{y}}_{ij}, \hat{\tilde{y}}_j^-)$, for all $i$ (18)

*3) Determine the ranking score*

Once the distances from ideal solutions are calculated for all alternatives, a ranking score ($RS_i$) can be formulated based on how much an alternative is close to the positive ideal solution and distant from the negative ideal solution as Eq. (19).

$RS_i = \frac{d_i^-}{d_i^- + d_i^+}$, for all $i$ (19)

The value of the ranking score is between 0 and 1. Bigger value for $RS_i$ denotes the higher rank for $i$th alternative.

## 5. Case Study

In this section, we implement the proposed MCGDM model on a real decision-making problem in the cyber-security area to demonstrate its applicability and effectiveness. A questionnaire survey was conducted on the difficulty of security components based on the judgement of a group of cyber-security experts [37, 38]. In this problem, a set of questions are used as criteria to rate 14 attack related security components ($N = 14$) by 38 experts ($K = 38$). The set of questions contains seven questions, which work as criteria ($M = 7$). Table 3 shows these questions and the derived criteria. Table 3 mentions whether a criterion is either a higher-the-better or a lower-the-better one, which are perceived from correlation with the overall difficulty in [38] and intuitively from the questions.

[Table 3]

Experts are asked to rate each component based on criteria using an interval between 0 and 100.

### 5.1. Implementation of the Proposed MCGDM

Once the data is collected, we can apply the proposed aggregation method to construct the decision cloud matrix. Table 4 shows the constructed decision cloud matrix.

[Table 4]

Now, we can determine the weights of criteria using the information in the decision cloud matrix by solving the bi-level optimisation model formulated in Eq. (12). The optimal weights are shown in Table 5. As can be seen, frequency has the highest weight, while interaction has the lowest weight.

[Table 5]

By determining the weights of criteria, we can construct the weighted decision cloud matrix. Table 6 shows the weighted constructed decision cloud matrix.

[Table 6]

To define positive and negative ideal solutions, we need to find the maximum and minimum cloud models for each criterion among the alternatives using Algorithm 4.

**Algorithm 4: Ideal solution finder**

**Input:** Set of cloud models for criterion $j$
**Output:** Indices of the maximum and minimum cloud models for criterion $j$ ($Max$ and $Min$)
1. Convert the cloud models into interval-values as described in *Definition 3*
2. $Max \leftarrow 1$ and $Min \leftarrow 1$
3. **for** $i$ from 2 to $N$ **do**
4.     Calculate $S_{\tilde{y}_{max}, \tilde{y}_i}$ /* Using Eq. (5) */
5.     Compare $\tilde{y}_{max}$ and $\tilde{y}_i$ /* Using $S_{\tilde{y}_{max}, \tilde{y}_i}$ and the rules in *Definition 3* */
6.     If $\tilde{y}_i > \tilde{y}_{max}$,
7.         $Max \leftarrow i$
8.     Then if $\tilde{y}_i < \tilde{y}_{max}$,
9.         Calculate $S_{\tilde{y}_{min}, \tilde{y}_i}$ /* Using Eq. (5) */
10.     Compare $\tilde{y}_{min}$ and $\tilde{y}_i$ /* Using $S_{\tilde{y}_{min}, \tilde{y}_i}$ and the rules in *Definition 3* */
11.     If $\tilde{y}_i < \tilde{y}_{min}$,
12.         $Min \leftarrow i$
13. **end for**

Considering whether each criterion is higher-the-better or lower-the-better (demonstrated in Table 3), the ideal solutions are defined as below.

$A^+$={(11.99,6.83,5.27), (6.01,10.97,9.68), (13.7,4.15,3.2), (4.33,9.69,7.98), (11.77,6.76,5.29), (11.95,5.29,4.02), (2.84,8.2,6.95)}

$A^-$={(6.97,8.59,6.83), (10.86,5.22,4.25), (3.59,7.5,5.9), (13.25,3.49,2.62), (5.59,10.14,8.4), (6.87,9.69,7.75), (8.05,9.18,7.27)}

The distances of each alternative from positive and negative ideal solutions can be calculated using Eqs. (17) and (18). Finally, these distance values are used to calculate the ranking scores which can be used to prioritise the alternatives. The values for distances and ranking scores as well as ranking of alternatives are brought in Table 7.

[Table 7]

As can be seen, the components $A_7$ and $A_9$ stand at the bottom, which means they are mostly vulnerable against the attack. On the other hand, the components $A_{12}$ and $A_6$ have the first and the second lowest levels in the ranking respectively, which denotes that they are in lower risk of being successfully attacked.

### 5.2. Comparative Analysis

In this section, we perform three comparative analyses on the ranking results based on the case study data. The first comparative analysis is a sensitivity analysis. A key parameter that has impact on the ranking of alternative in our proposed algorithm is the distance measure. Thus, we apply a sensitivity analysis on the distance measure. The second analysis is to compare the ranking performance of the proposed MCGDM algorithm with other group decision-making algorithms which use interval-valued data.

To identify the differences in the comparative analyses, we compute the ranking results and use a Spearman rank correlation test. This test is frequently used to determine the correlation between two sets of ranking values [39, 40]. The test statistic is measured using Eq. (20).

$$r_s = \frac{cov(R_X, R_Y)}{\sigma_{R_X} \sigma_{R_Y}}, \qquad (20)$$

where $R_X$ and $R_Y$ are two ranking vectors, $cov(.)$ is the covariance of two vectors, and $\sigma$ denotes the standard deviation of each vector. The value for $r_s$ is between -1 and 1 which denotes a complete negative and positive relationship between two ranking results, respectively. In other words, the higher the $r_s$ value, the more the similarity of two rankings.

The third analysis is to compare the computational time of the proposed MCGDM algorithm with the other group decision-making algorithms using interval values. In this analysis, we replicate the MCGDM procedure for each algorithm 100 times and record the CPU time. Then, compare the algorithms' computation using descriptive statistics and a statistical test.

*5.2.1. Sensitivity Analysis*

In this section, we analyse the sensitivity of the proposed MCGDM algorithm based on different distance measures. We consider two distance measures for cloud models previously introduced in the literature. One is proposed in [23] which treats each cloud model as a vector and finds the distance between two cloud models using Euclidean distance. Another one is proposed in [7] which is constructed based on Hamming distances.

The comparison of ranking results is shown in Fig. 8. For each distance measure, we calculate the Spearman test statistic. The test statistics for the ranking difference of the proposed distance measure with distance measures [23] and [7] are 0.891 and 0.767 respectively.

[Figure 8]

As can be seen in the ranking results, the components $A_{12}$, $A_6$, $A_3$ and $A_2$ are ranked high and the components $A_7$ and $A_9$ rank at the bottom across different similarity measures. This show that the proposed MCGDM algorithm is robust in terms of using distance measures. Moreover, the Spearman rank-correlation statistics confirm this robustness. They denote that the ranking results from different similarity measures are strongly correlated and similar to the ranking results of the proposed MCGDM.

*5.2.2. Ranking Performance Comparison*

As previously mentioned, we compare our MCGDM algorithm with two other group decision-making algorithms [21, 41], which used interval-valued data as inputs, using the case study.

In [41], the interval-valued data are firstly normalized. Then, the normalized data across decision-makers are aggregated into interval-valued intuitionistic fuzzy numbers (IVIFNs) using the mean and standard deviation values of the endpoints. so, there would be a decision matrix which the values are IVIFNs. After that, the weighted average IVIFN for each alternative is calculated. Finally, the alternatives are ranked based on the calculated weighted average.

In [21], the interval-valued data (with uniform membership function) collected from decision-makers are aggregated into types-2 fuzzy sets (as described in section 2.1). Each FS is de-fuzzified by obtaining the centroid. The weighted average of the centroid of the criteria is used to prioritise the alternatives. Fig. 9 provides a comparison between ranking results of the algorithms. The test statistics for the ranking difference of the proposed MCGDM algorithm with algorithms [41] and [21] are 0.621 and 0.942 respectively.

[Figure 9]

The Spearman correlation statistics are different between the comparisons. In comparison with [41], the test statistic is low and the alternatives which are ranked at the top or bottom vary. Whereas, in comparison with [21], the correlation statistic is high, and also the components $A_{12}$, $A_3$ and $A_2$ are ranked at the top and the components $A_7$ and $A_9$ the ranked at the bottom in both rankings. So, the ranking results of the proposed algorithm is similar to [21] and is distant from [41].

By more investigations, We find out that there are several deficiency and drawback in the algorithm of [41]. We believe these drawbacks to be as follows. (1) In the normalization step, if the minimum value of the lower bounds of intervals is zero, the interval values are transformed into [0, 1] intervals. This transformation leads to the loss of a lot of information. (2) In the aggregation process, the assumption is that the mean value of the lower bounds for the normalized intervals (for each criterion and each alternative) is less than 0.5, which is not always true in reality including in the case study. So, the authors formulated the aggregation in a way to control that the mean plus standard deviation (for lower bounds) does not exceed 0.5. However, based on the previous assumption, they did not control the value of mean minus standard deviation. This value exceeds 0.5 several times in the aggregating decision-makers' ratings, which leads to having an upper bound lower than the lower bound. In our implementation, we fixed the problem by applying a similar control. (3) There are similar assumptions and controls for the upper bound values, so that the mean value of upper bounds for the normalized intervals (for each criterion and each alternative) should be greater than 0.5. (4) Applying these controls lead to losing information during aggregation.

On the other hand, the algorithm presented in [21] is valid for interval-valued data with uniform membership function. So, having similar ranking results with [21] and different ranking results from [41] is a validation evidence for our proposed MCGDM algorithm.

*5.2.3. Computational Time Comparison*

The statistics for computation time (in milliseconds) of each MCGDM algorithm after 100 times of running are presented in Table 8.

[Table 8]

The statistics show that the proposed MCGDM needs less CPU time for computations in average. Also, to check the significance of the difference among the algorithms, we used Kruskal-Wallis test which is a non-parametric test (as CPU time data records do not have Normal distribution). The test proves that computation times of the algorithms are significantly different.

Overall, there are advantages for the proposed MCGDM algorithm in comparison with two others as follows.

1. It does not have the deficiencies mentioned for [41] and its performance is validated through the comparative analysis.
2. Our algorithm has a built-in mechanism to find the optimal weights for criteria objectively and automatically.
3. The aggregation method is formulated in a way that needs less computations, to gain the information from the decision-makers.

## 6. Conclusion

In this paper, we propose a MCGDM algorithm under uncertainty with interval-valued data. In this algorithm, we propose a new aggregation method using cloud models and an extended version of TOPSIS method. The algorithm has an automated mechanism to derive the weights of criteria using the proposed bi-level optimisation model. This paper contributes to the MCGDM literature as listed below.

- The proposed method gives more flexibility to decision-makers to express the uncertainty in their opinions.
- Using cloud models for aggregation provides a simpler representation and therefore fewer computations and less memory usage.
- There is no need for more inputs from decision-makers to weight criteria.

We design an experimental setup to examine how well the aggregation method can retain the information from decision-makers. Also, we used the experimental setup to compare the performance of the aggregation method with other methods.

Moreover, the algorithm is applied on a practical case of a cyber-security problem to prioritise security components based on their vulnerability against cyber-attacks. A comparative analysis of the ranking results for the proposed MCGDM algorithm is undertaken. The comparative analysis includes sensitivity analysis and comparisons with other existing algorithms.

The results show that the aggregation method is robust over various numbers of decision-makers (experimented in section 3.1 and results presented in Fig. 5). Also, it performs well and retains information in comparison with other comparable methods (experimented in section 3.1 and results presented in Fig. 6) and performs better than other methods gaining information from interval-valued data with Gaussian membership function (experimented in section 3.1 and results presented in Fig. 7). The proposed MCGDM algorithm is robust in terms of distance measures (experimented in section 5.2.1 and results presented in Fig. 8). The ranking results of the algorithm are valid in comparison with the existing algorithms (experimented in section 5.2.2 and results presented in Fig. 9). The proposed MCGDM algorithm needs less computation time in comparison with the existing ones (experimented in section 5.2.3 and results presented in Table. 8).

The assumption of Gaussian membership functions for intervals could be considered as a limitation of the proposed method. Even though we justified that this membership function is the most appropriate one for intervals, we suggest considering other membership functions or a combination of various membership functions for future research, . So, new aggregation methods should be developed accordingly. The performance of the aggregation methods can be examined using the proposed experimental design.

Moreover, the proposed aggregation method can be used for ensemble learning using machine learning methods by considering the uncertainty level in their prediction. Furthermore, the proposed algorithm can be applied to various types of collected uncertain data other than expert opinions. As the interval-valued data for this algorithm are assumed to have Gaussian membership functions, the algorithm fits well to uncertain data which is sourced from measurement errors, learned prediction errors and random observations. The algorithm can be used for consensus problems when the experts are not humans but are intelligent machines or expert systems.